\documentclass[sigconf]{acmart}
\setcopyright{none}
\renewcommand\footnotetextcopyrightpermission[1]{}
\usepackage{xcolor}
\definecolor{JazzyBlue}{RGB}{89,56,212}
\definecolor{GreenBlue}{RGB}{13,152,186}
\definecolor{MurkeyBlue}{RGB}{90,96,134}

\AtBeginDocument{%
  }

\copyrightyear{2023}
\acmYear{2023}
\setcopyright{acmlicensed}\acmConference[MM '23]{Proceedings of the 31st ACM International Conference on Multimedia}{October 29-November 3, 2023}{Ottawa, ON, Canada}

\acmBooktitle{Proceedings of the 31st ACM International Conference on Multimedia (MM '23), October 29-November 3, 2023, Ottawa, ON, Canada}
\acmPrice{15.00}
\acmDOI{10.1145/3581783.3612139}
\acmISBN{979-8-4007-0108-5/23/10}

\begin{document}

\title{Multimodal Color Recommendation in Vector Graphic Documents}

\author{Qianru Qiu}
\email{qiu\_qianru@cyberagent.co.jp}
\affiliation{%
  \institution{CyberAgent Inc.}
  \streetaddress{2-24-12 Shibuya}
  \city{Shibuya}
  \state{Tokyo}
  \country{Japan}
  \postcode{150-6121}
}

\author{Xueting Wang}
\email{wang\_xueting@cyberagent.co.jp}
\affiliation{%
  \institution{CyberAgent Inc.}
  \streetaddress{2-24-12 Shibuya}
  \city{Shibuya}
  \state{Tokyo}
  \country{Japan}
  \postcode{150-6121}
}

\author{Mayu Otani}
\email{otani_mayu@cyberagent.co.jp}
\affiliation{%
  \institution{CyberAgent Inc.}
  \streetaddress{2-24-12 Shibuya}
  \city{Shibuya}
  \state{Tokyo}
  \country{Japan}
  \postcode{150-6121}
}


\begin{abstract}
  Color selection plays a critical role in graphic document design and requires sufficient consideration of various contexts. However, recommending appropriate colors which harmonize with the other colors and textual contexts in documents is a challenging task, even for experienced designers. In this study, we propose a multimodal masked color model that integrates both color and textual contexts to provide text-aware color recommendation for graphic documents. Our proposed model comprises self-attention networks to capture the relationships between colors in multiple palettes, and cross-attention networks that incorporate both color and CLIP-based text representations. Our proposed method primarily focuses on color palette completion, which recommends colors based on the given colors and text. Additionally, it is applicable for another color recommendation task, full palette generation, which generates a complete color palette corresponding to the given text. Experimental results demonstrate that our proposed approach surpasses previous color palette completion methods on accuracy, color distribution, and user experience, as well as full palette generation methods concerning color diversity and similarity to the ground truth palettes.
\end{abstract}


\begin{teaserfigure}
  \includegraphics[width=\textwidth]{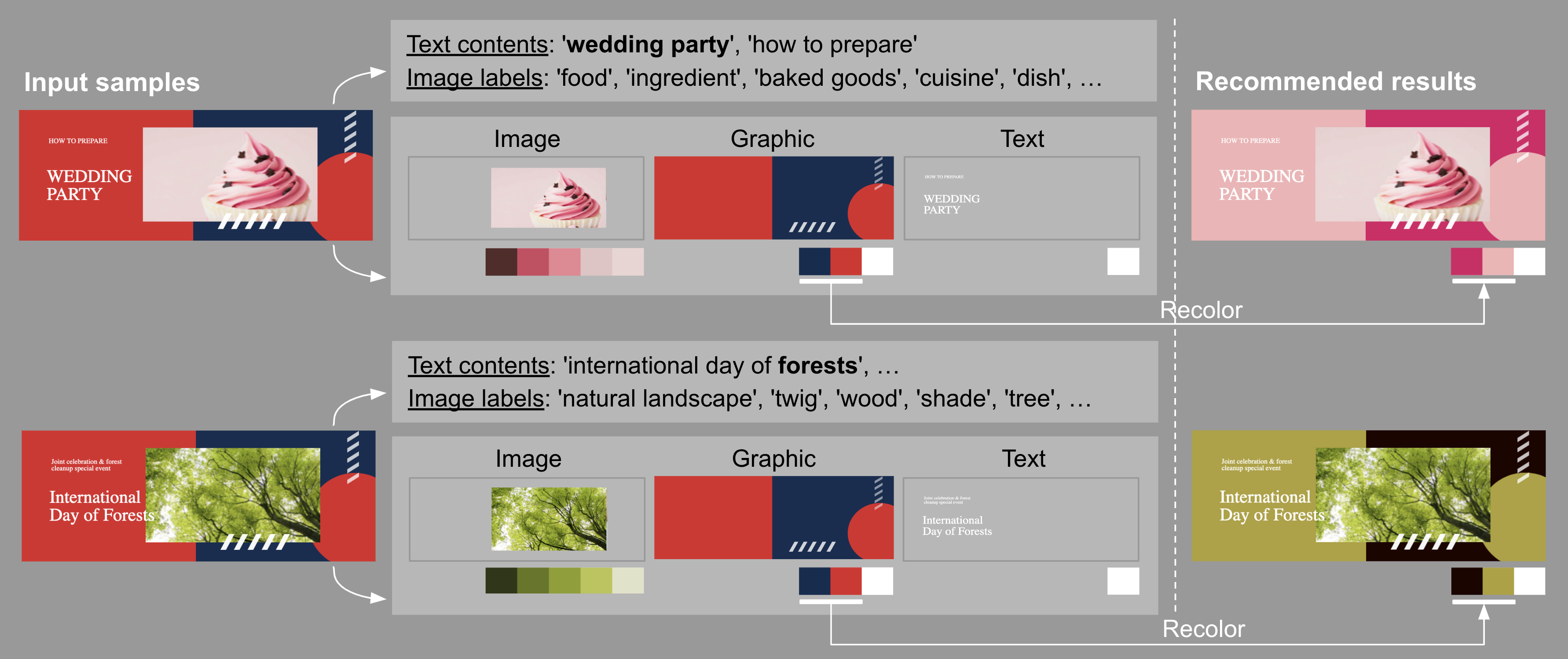}
  \caption{Concept of multimodal color recommendation in vector graphic documents. The design samples on the left have textural information in the form of text contents and image labels. We extract color palettes for each visual element, such as image, graphic, and text, and reorder the colors based on color lightness. Our method replaces specified colors in the palettes with new colors based on the textual information and surrounding colors. By recoloring the original ones with the new colors, we obtain variants of the designs. The elements in the input samples are from the Crello-v2 dataset.}
  \label{fig:overview}
\end{teaserfigure}


\maketitle

\section{Introduction}
In graphic design, many creative applications offer thousands of templates that provide a layout for the design. The graphic design workflow begins with the selection of design templates, followed by the replacement of specific design elements, such as images and text contents. After customizing the partial design elements, as the input samples in Figure~\ref{fig:overview}, designers often adjust colors for the specific elements to ensure harmony of the whole colors and alignment with the semantic concepts in the design.
Semantic concepts are often represented by text and image contents.
For example, in the bottom row of Figure~\ref{fig:overview}, a promotional poster displays a headline of `International Day of Forests' and an image of trees featuring a green color palette. 
If we only consider color harmony theories~\cite{chijiiwa1987color}, various color combinations are possible for the graphic elements to match the green colors in the image, such as monochromatic colors with various tones of green, complementary colors like yellow-green, or analogous colors like blue-green. Designers might choose various tones of green for the graphic elements to maintain consistency with the nature-related concept. 
To achieve a cohesive and harmonious outcome, it is crucial to consider both semantic and color aspects in the design process. 

A color palette refers to a specific set of colors in refined forms. A well-chosen color palette helps communicate the intended message effectively. In this study, the primary objective of color recommendation is color palette completion, which means suggesting appropriate colors in palettes, taking into consideration both the existing color and text contexts. In addition, we address the full palette generation task, which is a specific case of color recommendation. This task focuses on creating a set of harmonious colors based only on the given textual contexts, without factoring in any other color context. In this study, we present a versatile model that can effectively perform both color recommendation tasks.

Color palettes in vector graphic documents are complex due to the presence of multimodal contents, including images, shapes, and texts. In recent years, data-driven deep learning techniques have demonstrated potential in color palette recommendation for graphic documents. A recent study \cite{qiu2023color} introduced a masked color recommendation model for graphic documents. However, this work only examined the relationships among colors in multiple palettes.
Some studies based on multi-modality learning have aimed to generate a color palette based on textual information for image colorization \cite{bahng2018coloring, maheshwari2021generating}. The text in these works comprises a brief sequence, such as a single word (e.g., `sunny'), an attribute-object pair (e.g., `cute dog'), or a phrase (e.g., `grape to strawberry'). They employed a generator with a text encoder to represent the input text.
However, the small scale of training data restricted the capacity to encode and represent complex textual information. In contrast, using text embeddings generated by large language models (LLMs) is more comprehensive than learning from the small-scale dataset.

In this paper, we propose a multimodal masked color model for text-aware color recommendation in graphic documents using the designed attention networks. We utilize the pre-trained CLIP model to obtain comprehensive text embeddings that can represent both textual and visual features, enabling us to consider a broader range of text representation for color recommendation. We primarily conduct a series of quantitative and qualitative evaluations on color palette completion for graphic documents. In addition, we present experiments on full palette generation to validate the versatility and effectiveness of our proposed method.

The main contributions of this paper can be summarized as:
\begin{itemize}
\setlength{\parskip}{0cm}
\setlength{\itemsep}{2mm}
    \item We propose a multimodal masked color model to incorporate color and textual contexts within a graphic document using CLIP-based text representation and designed integration networks.
    \item Our proposal is applicable for two color recommendation tasks: color palette completion and full palette generation.
    \item Our proposed method outperformed the state-of-the-art methods on color palette completion task for graphic documents on accuracy and user experience. 
    \item Our proposed method surpassed previous methods on full palette generation task on color diversity and similarity to the ground truth palettes.
\end{itemize}

\section{Related works}
Our approach is applicable for two scenarios: color palette completion, which recommends colors for the given palettes, and full palette generation, which generates a complete color palette in accordance with specified themes or semantic requirements. In this section, we review the major related works on color palette completion and full palette generation.

\subsection{Color palette completion}
In general, recommending colors based on the given colors relies on the color harmony of color combinations and human aesthetic preferences. O’Donovan \textit{et al.}~\cite{o2011color} and Kita \textit{et al.}~\cite{kita2016aesthetic} proposed to expand one or more new colors for a given set of colors. These prior researches depend on hand-crafted feature extraction methods and train regression models to learn the weight of each feature. These features include palette colors, mean, standard deviation, median, max, and min across a single channel in different color spaces. However, these hand-crafted features may not fully capture the semantics of colors, and some features may not significantly affect color prediction.

Recently, some researchers have explored deep learning algorithms for color recommendation in various design fields. Kim \textit{et al.}~\cite{kim2022colorbo} employed an extension of the Word2Vec model to train a color embedding model, which recommended additional colors based on a series of given colors for Mandala Coloring. Yuan \textit{et al.}~\cite{yuan2021infocolorizer} employed a Variational AutoEncoder with Arbitrary Conditioning (VAEAC) model to interactively recommend colors for elements in infographics, e.g., shapes, pictograms, and text. Qiu \textit{et al.}~\cite{qiu2022intelligent} proposed a Transformer-based masked color model to recommend specific region colors on landing pages. Most recently, Qiu \textit{et al.}~\cite{qiu2023color} proposed an improved Transformer-based masked color model for vector graphic documents. This model utilizes multiple color palettes extracted from multiple design elements in the Crello dataset~\cite{yamaguchi2021canvasvae} and recommends colors corresponding to the existing ones in multiple palettes. However, the accuracy of recommended colors is relatively limited, and the recommended results exhibit significantly worse performance than the ground truth in user studies.

\subsection{Full palette generation}
Many web-based services, such as COLOURLovers~\cite{ColourLovers}, and Adobe Color~\cite{AdobeColor}, offer color palette templates created by users or extracted from images. These palettes are categorized using various semantic tags, such as `fashion', `love and heart'. Some early studies in magazine cover design~\cite{jahanian2013recommendation, yang2016automatic} used predefined palettes for design elements based on particular themes.
Subsequent research has introduced neural network architectures to predict a single color for text~\cite{kawakami2016character, monroe2017colors}. Kikuchi \textit{et al.}~\cite{kikuchi2023generative} proposed maximum likelihood estimation (MLE) and conditional variational autoencoder (CVAE) models using the Transformer-based network to recommend text and background colors for each element with text content in e-commerce mobile web pages. Moreover, Maheshwari \textit{et al.}~\cite{maheshwari2021generating} proposed a conditional generative adversarial networks (GAN) architecture for generating a color palette for image colorization with an attribute-object pair text input, such as `warm sunshine' and `cute dog'. Likewise, Bahng \textit{et al.}~\cite{bahng2018coloring} proposed a conditional GAN based text-to-palette generation networks for image colorization that reflect the semantics of text input. They also introduced a Palette-And-Text (PAT) dataset. 

Overall, the study of deep learning based color recommendation is in the preliminary stage, and the recommended results are still far from the actual designs made by designers. There is also no unified method that has been applied to both color palette completion and full palette generation tasks. Therefore, we propose a versatile multimodal masked color model and explore improving the model performance for both color recommendation tasks.

\section{Approach}
\begin{figure}[t]
    \centering
    \includegraphics[width=1.0\linewidth]{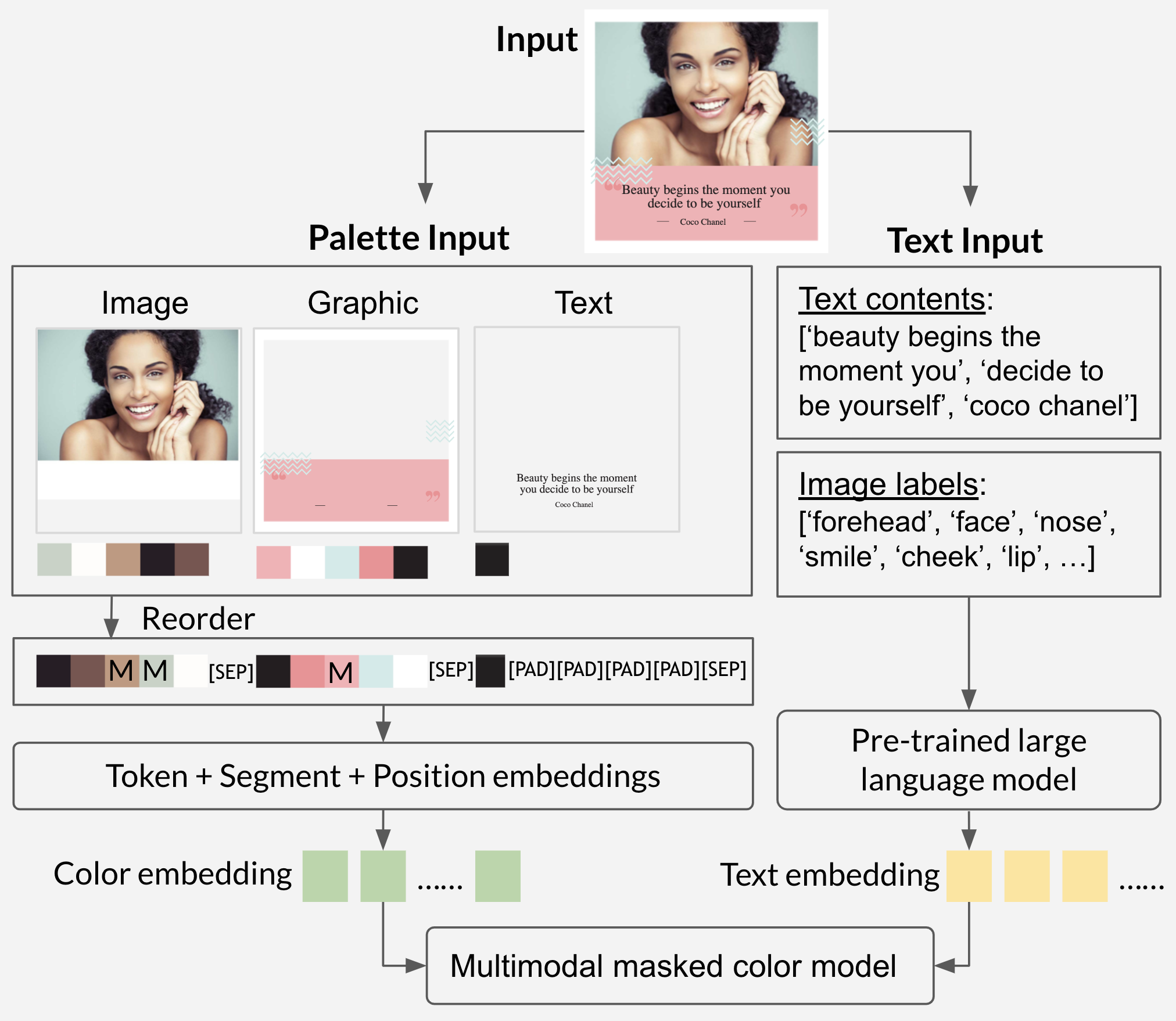}
    \caption{Extraction and representation processes of color and text in a graphic document. The palettes are extracted from the multiple visual elements, such as image, graphic, and text. The text input contains text contents and image labels.}
    \label{fig:representation}
\end{figure}

The elements in graphic documents can be classified into three primary groups: image, graphic, and text, as shown in Figure~\ref{fig:representation}. Graphics encompass decorative elements such as lines, shapes, and other vector-based illustrations. In this paper, we extract multiple color palettes from these three element groups and reorder the color positions. Additionally, we extract varied textual contexts in image and text elements, and obtain text embeddings using pre-trained LLMs. The overview of our framework is illustrated in Figure~\ref{fig:representation}. We apply a multimodal masked color model to integrate the color and text representations for color sequence completion and predict the specified colors with a high probability for color recommendation.

\subsection{Reordered color representation}
 We adopt the color extraction method in the related work by Qiu \textit{et al.}~\cite{qiu2023color}. They ordered colors in each palette according to the size of their color clusters, reflecting their color area size. However, we found that area-based color ordering does not positively impact model performance. One possible explanation is that the size of a color area is not necessarily proportional to its importance or relevance in the overall color context. For example, a small but prominent red color in a predominantly green design may have a greater impact on the color context than a large but subtle green color. On the other hand, lightness provides a simple and intuitive way to organize colors. It is commonly used for color lexicographical sorting~\cite{zaccarin1993novel} and has been applied to various applications, especially for image recoloring~\cite{chang2015palette, zhang2017palette, cho2017palettenet}. The gradient lightness feature has been identified as an important factor for palette aesthetic ratings~\cite{o2011color}. Consequently, in this approach, we utilize lightness in CIELAB color space as the basis for color order, which we believe will provide a richer color context.

 In our color model, a color is signified as a word, and a palette is signified as a sentence. Color representation can be divided into three primary stages: color quantization, encoding quantized colors into learnable embeddings, and training with a masked autoencoder approach. Color quantization aims to reduce the number of colors. In this work, RGB color data is first converted to CIELAB color space, with a range of [0, 255], and each color is assigned to one of the bins in a $b\times b\times b$ histogram, with $b=16$, labeled as a color code. Then quantized color codes are encoded into vectors and embedded in the space during the learning progress. For palettes shorter than the fixed maximum length, we add the [PAD] token to complete the length. Additionally, the [SEP] token is appended at the end of a palette. The palettes of image, graphic, and text elements, are respectively labeled with the segment number 1, 2, and 3. Given a sequence of multiple palettes, we construct its input representation by summing the corresponding token embeddings, segment embeddings, and position embeddings. Once the learnable embeddings for colors are obtained, the training process can begin with masking some percentage of the input tokens at random with [MASK] token, and subsequently predicting those masked tokens.

\subsection{Text representation}
Textual information plays a crucial role in conveying semantic concepts and achieving design goals. Color variations often rely on figurative language~\cite{kawakami2016character}, such as \textcolor{MurkeyBlue}{murkey blue}, \textcolor{GreenBlue}{greeny blue}, \textcolor{JazzyBlue}{jazzy blue} (best viewed
in color). By incorporating textual information into the design process, models can recommend colors that are more contextually relevant and harmonious with the intended concept, resulting in more effective design outcomes. Design works have various forms of textual information. In this study, we gather two types of text from the graphic documents: text contents and image labels, as demonstrated in Figure~\ref{fig:representation}. Text contents are obtained directly from the original text elements. Image labels are extracted from image elements using object detection techniques. In our work, we utilize Google Vision API to detect multiple objects in images and rank the label words by their confidence scores. 

In recent years, pre-trained LLMs have made significant advancements in natural language processing (NLP). These models have exhibited remarkable results across various applications. In this study, we use the CLIP~\cite{radford2021learning} model to obtain the text embeddings of each text content and image label. CLIP is pre-trained on a large corpus of images and their associated captions, which allows it to generate embeddings that capture both textual and visual information. The colors present in the input image can be correlated with the textual information during the training process, and we believe that this connection has the potential to improve the performance of text-aware color models. We employ the CLIP model to obtain text embeddings for each text content and image label. In addition, we compare the performance of CLIP-based text embedding with that of BERT-based text embedding to evaluate the impact of the chosen LLMs on the results. The finding of this comparison will be discussed in the experiment section. 

\subsection{Multimodal masked color model}

\begin{figure}[t]
    \centering
    \includegraphics[width=1.0\linewidth]{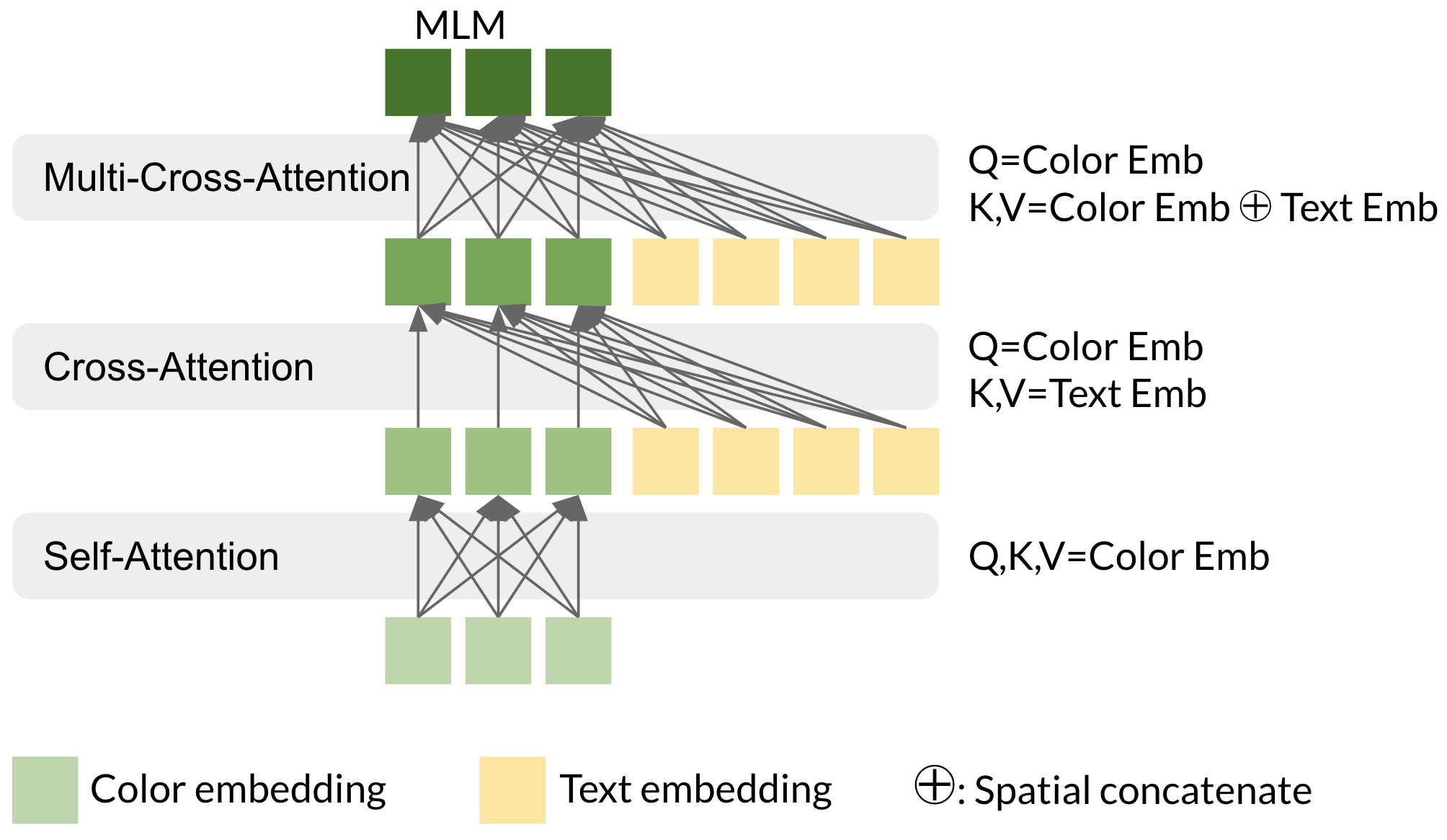}
    \caption{Multimodal masked color model including one self-attention and two cross-attention networks with color and text input representations.}
    \label{fig:model}
\end{figure}

As shown in Figure~\ref{fig:model}, our proposed multimodal masked color model is composed of one self-attention module and two cross-attention modules, designed to effectively integrate color and textual contexts. An attention module is described as mapping a query (Q) and a set of key (K) and value (V) pairs to an output~\cite{vaswani2017attention}. In the self-attention module, we have input color embeddings $c = \{c_1, c_2, ..., c_N\}$, where $N$ is the number of colors. The output $O_{sa}$ of the self-attention is computed by attention function with $Q_{sa} = K_{sa} = V_{sa} = c$. This module can effectively capture the intra-relationship among colors within the same palette and the inter-relationship between different palettes. To model the inter-relationship between colors and text, we introduce cross-attention modules. In the cross-attention module, we have color embeddings from the self-attention module as queries, while keys and values correspond to text embeddings $t = \{t_1, t_2, ..., t_M\}$, where $M$ is the number of textual phrases. Thus, the output $O_{ca}$ of the cross-attention is computed with $Q_{ca} = O_{sa}$ and $K_{ca} = V_{ca} = t$. Additionally, the multi-cross-attention module is a comprehensive network that captures both intra-relationship and inter-relationship in colors and text. In the multi-cross-attention module, the queries are color embeddings from the cross-attention module, while keys and values are obtained by concatenating color embeddings from the cross-attention layer and text embeddings. Thus, the output $O_{mca}$ of the multi-cross-attention is computed with $Q_{mca} = O_{ca}$ and $K_{mca} = V_{mca} = \text{concat}(O_{ca}, t)$, where $\textrm{concat}(\cdot, \cdot)$ denotes the concatenation operation.

The masked color model following a methodology similar to a masked language model (MLM)~\cite{devlin2018bert}, randomly masks the color tokens from the input, and then predicts the masked tokens based on their contexts. In an MLM, the masking rate refers to the proportion of tokens that are masked in the input sequence, and masking strategies determine how the tokens are selected for masking. In this study, color tokens are randomly chosen for masking based on the masking rate, and are then replaced with the [MASK] token based on the masked token rate. The masking rate and the masked token rate have a significant impact on model performance as reported in~\cite{wettig2022should}. 

Utilizing a self-supervised learning approach, the model can discern the underlying connections between colors and textual contexts present in existing designs. During the training process, we compute the cross-entropy loss between the actual colors and the predicted ones. 
Once the color model is trained, it can then be utilized to predict specific colors with a high probability and generate color palettes that harmonize with textual contexts.

\section{Experiments of color completion}
To demonstrate the effectiveness of our proposed model, we conducted quantitative and qualitative experiments on the Crello-v2 dataset \cite{yamaguchi2021canvasvae}. We primarily conducted a comparison on the color palette completion task between our proposed multimodal masked color model with the related work by Qiu \textit{et al.} \cite{qiu2023color}, which only incorporates color representation in its masked model. To evaluate model performance, we measured the accuracy metric by comparing the predicted colors with the ground-truth colors. Since black and white colors are frequently present in design data, it is essential to ensure that the improvement in color prediction accuracy does not disproportionately focus on these two colors. To address this issue, we evaluated the distribution of the correctly predicted colors by calculating the Shannon entropy. A low distribution value signifies that the model predominantly predicts a limited number of colors with high frequency, while a higher distribution value indicates that the model predicts a broader range of colors with more balanced frequencies. The color distribution value serves as a measure to quantify the diversity of the correctly predicted colors. It is worth noting that incorrect predictions may increase color diversity but decrease palette quality~\cite{bahng2018coloring}. Therefore, these outcomes are not taken into account in this experiment. In addition, we compare the recommended colors with the highest probability of each model, denoted as `@1' in the subsequent results.

\subsection{Datasets}
The Crello-v2 dataset provides complete document structure and element attributes, including the type of the element, position, size, opacity, text contents, or a raster image. We collected the Multi-Palette-And-Text dataset compiled from the Crello-v2 dataset. It contains multiple palettes for image, graphic, and text elements, and textual information of text contents and image labels. The resulting dataset comprises 14,020 / 1,704 / 1,712 valid data for training, validation, and testing, all of which contain image-graphic-text palettes and English text contents. Each palette contains up to five colors and each text sequence contains up to ten phrases in the experiments.

\subsection{Implementation details}
In all models of this experiment, each color and text token is transformed into a 512-dimensional vector. The self-attention module employs three attention layers with eight heads. The two cross-attention modules are used with one attention layer and one head. All models are trained using early stopping with a patience of 30, which stops the training process after 30 epochs of no improvement in the validation loss. To obtain reliable results, we conducted training for each model across five runs and calculated the mean value of accuracy and distribution.
During training with the lightness-based ordered color dataset, we randomly choose {40\%} of the color token positions for prediction, and replace the token with the [MASK] token {50\%} of the time. On the other hand, a masking rate of {30\%} and a masked token rate of {30\%} are found to yield relatively optimal model performance for the area-based ordered color dataset. We used a NVIDIA T4 GPU and a training process was completed within an hour.

\subsection{Performance comparison}
\begin{figure}[t]
    \centering
    \includegraphics[width=1.0\linewidth]{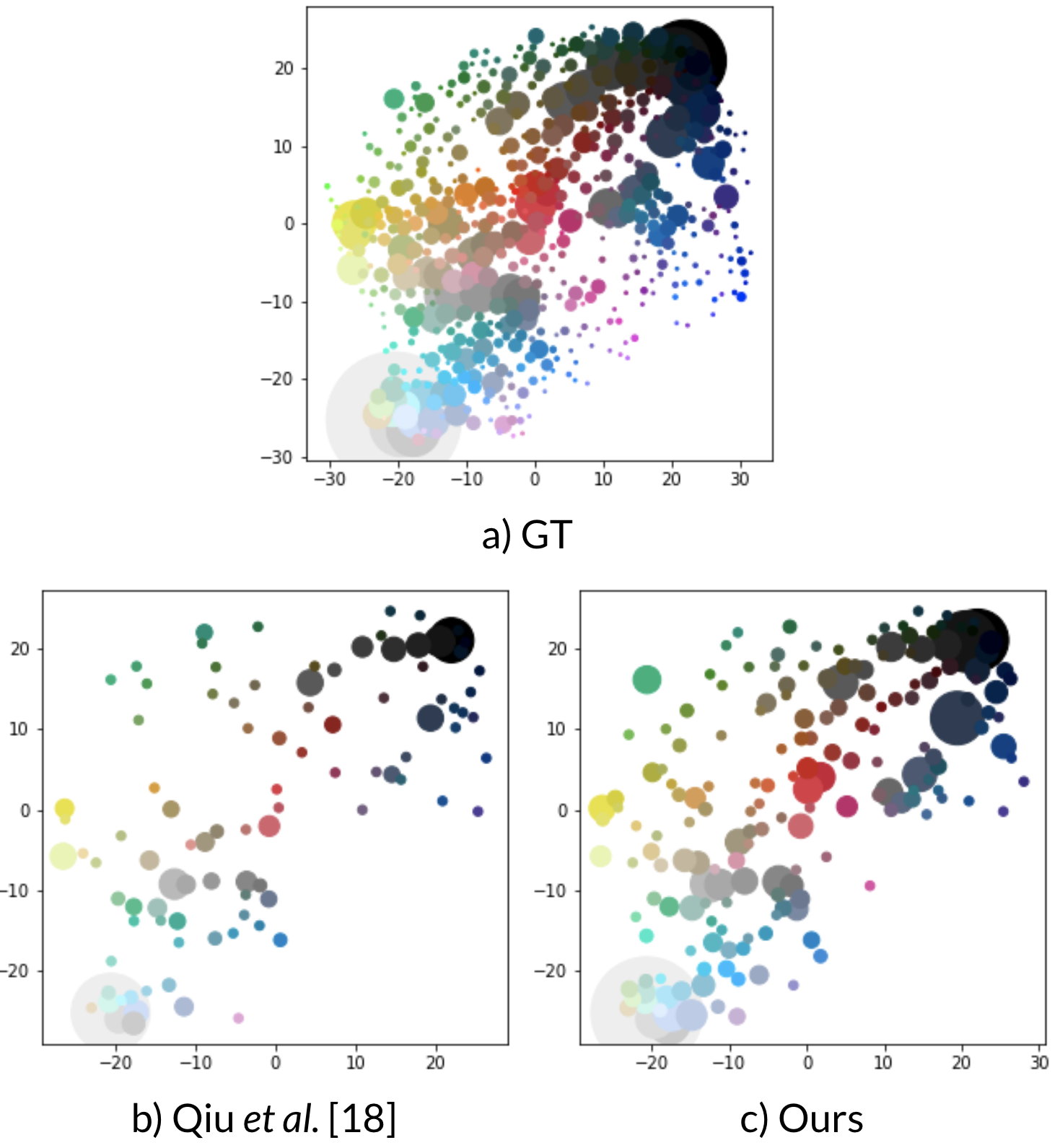}
    \caption{Color distributions that the data point is assigned with its own color and the point size reflects its frequency. a) displays all the ground truth colors in the multiple palettes of the test dataset. b) displays the correctly predicted three-color results of the method by Qiu \textit{et al.}~\cite{qiu2023color}. c) displays the correctly predicted three-color results of our method.}
    \label{fig:color_distr}
\end{figure}

The comparison results of our proposed method and the related works are shown in Table~\ref{tab:overall_res}. Our results indicated that our proposed method has higher accuracy than the method by Qiu \textit{et al.}~\cite{qiu2023color} for predicting different numbers of masked colors. Furthermore, the higher distribution value of our results demonstrated that the correctly predicted colors have a broader range of colors with balanced frequencies. 
To visualize the distribution of the correctly predicted colors, we utilized the 3-dimensional CIELAB color data and transform it into 2-dimensional space by t-SNE (t-Distributed Stochastic Neighbor Embedding). We selected each best-trained model with the highest validation accuracy and output the results for three-color prediction based on the test dataset. The distribution results are shown in Figure~\ref{fig:color_distr}. It demonstrated that the output of our model has denser points, indicating that our model had more accurate predictions, and the accurately predicted colors were not limited to black and white colors. 

\begin{table}[t] 
    \caption{Comparison results of accuracy@1 and distribution@1 for predicting different numbers of masked colors.} 
    \label{tab:overall_res} 
    \setlength\tabcolsep{0pt} 
    \resizebox{\linewidth}{!}
    {
        \begin{tabular}{@{\extracolsep{8pt}} lcccccc}
        \hline
        \toprule
        & \multicolumn{3}{c}{Accuracy@1$\uparrow$} & \multicolumn{3}{c}{Distribution@1$\uparrow$}\\
        \cmidrule{2-4} \cmidrule{5-7}
        Method & {1 color} & {2 colors} & {3 colors} & {1 color} & {2 colors} & {3 colors}\\ 
        \midrule
        Qiu \textit{et al.}~\cite{qiu2023color} & 36.72\% & 16.04\% & 6.45\% & 4.93 & 4.31 & 2.94 \\ 
        Ours & \textbf{47.13\%} & \textbf{26.22\%} & \textbf{15.67\%} & \textbf{6.04} & \textbf{6.64} & \textbf{6.30} \\
        \bottomrule
        \end{tabular}
    }
\end{table}

\subsection{User study}

\begin{figure*}[t]
    \centering
    \includegraphics[width=1.0\textwidth]{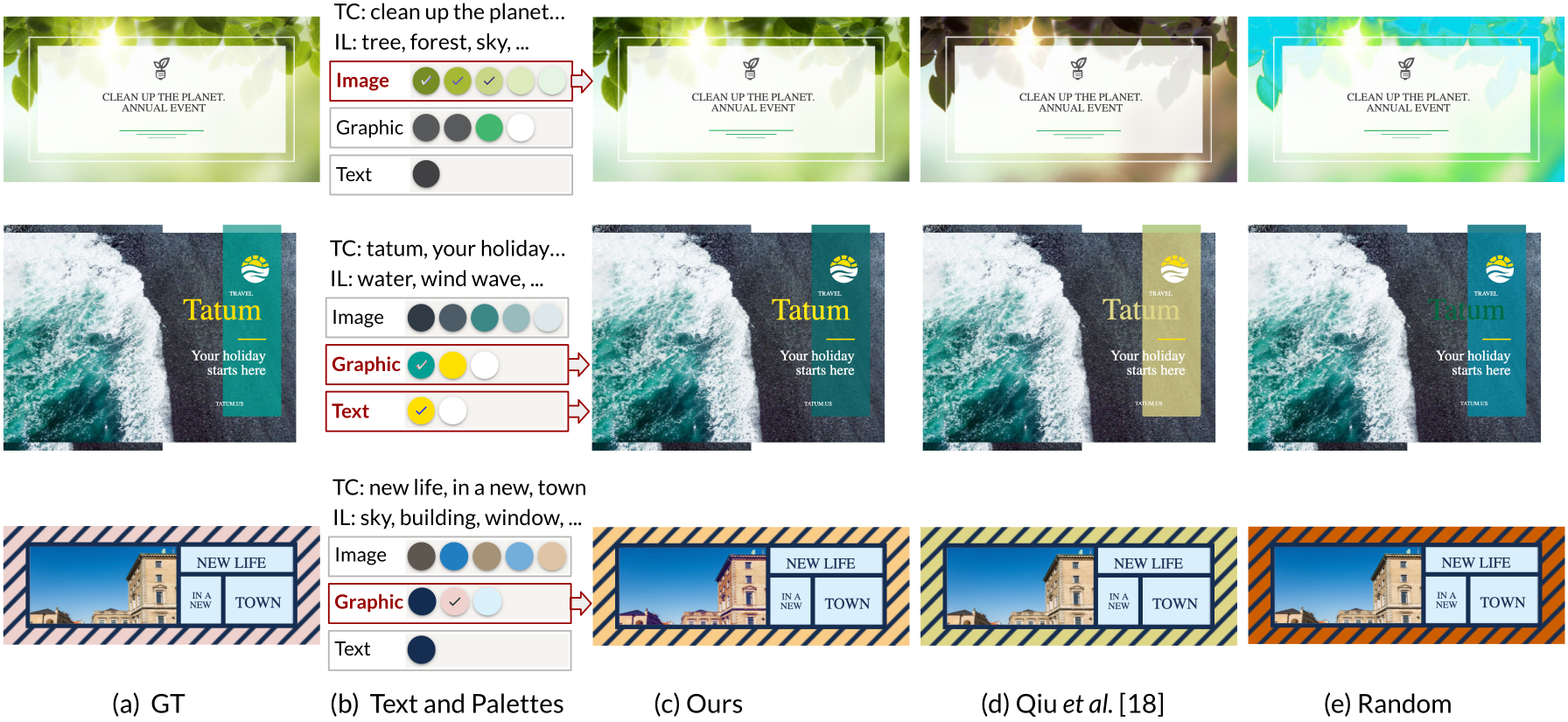}
    \caption{Color recommendation results with our proposed method, Qiu \textit{et al.}~\cite{qiu2023color}, and random color selection. The selected colors for recoloring are marked with `\checkmark'. In the first sample, three colors in image element are recolored. In the second sample, two colors are recolored: one in graphic element and the other in text element. In the third sample, one color in graphic element is recolored. (b) Text and Palettes are extracted from GT, including text contents (TC), image labels (IL), and the palettes of image, graphic, and text elements.}
    \label{fig:recolor_res}
\end{figure*}

\begin{figure}[t]
    \centering
    \includegraphics[width=1.0\linewidth]{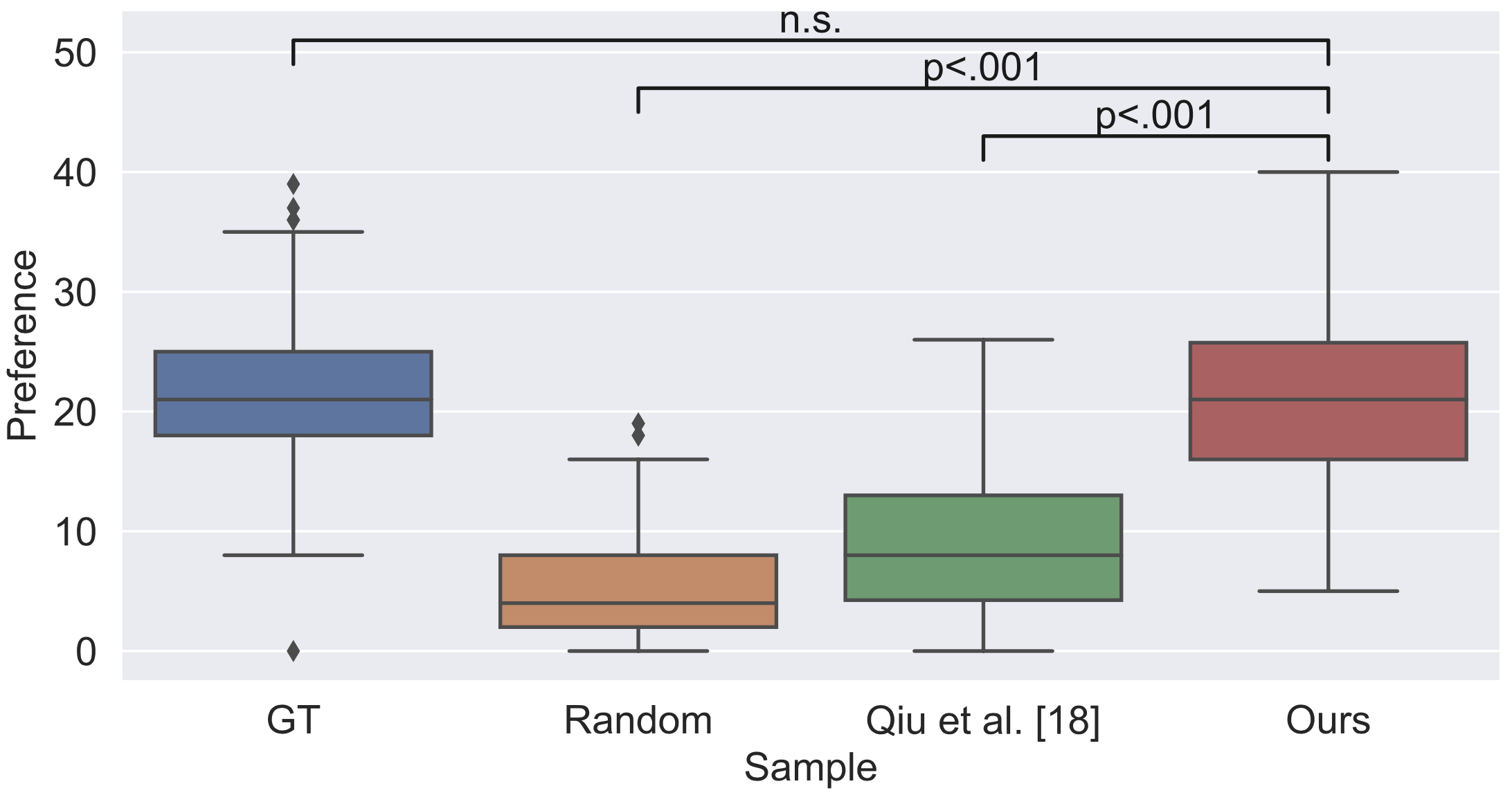}
    \caption{Evaluation results of good design from designers.}
    \label{fig:eval_good}
\end{figure}

\begin{figure}[t]
    \centering
    \includegraphics[width=1.0\linewidth]{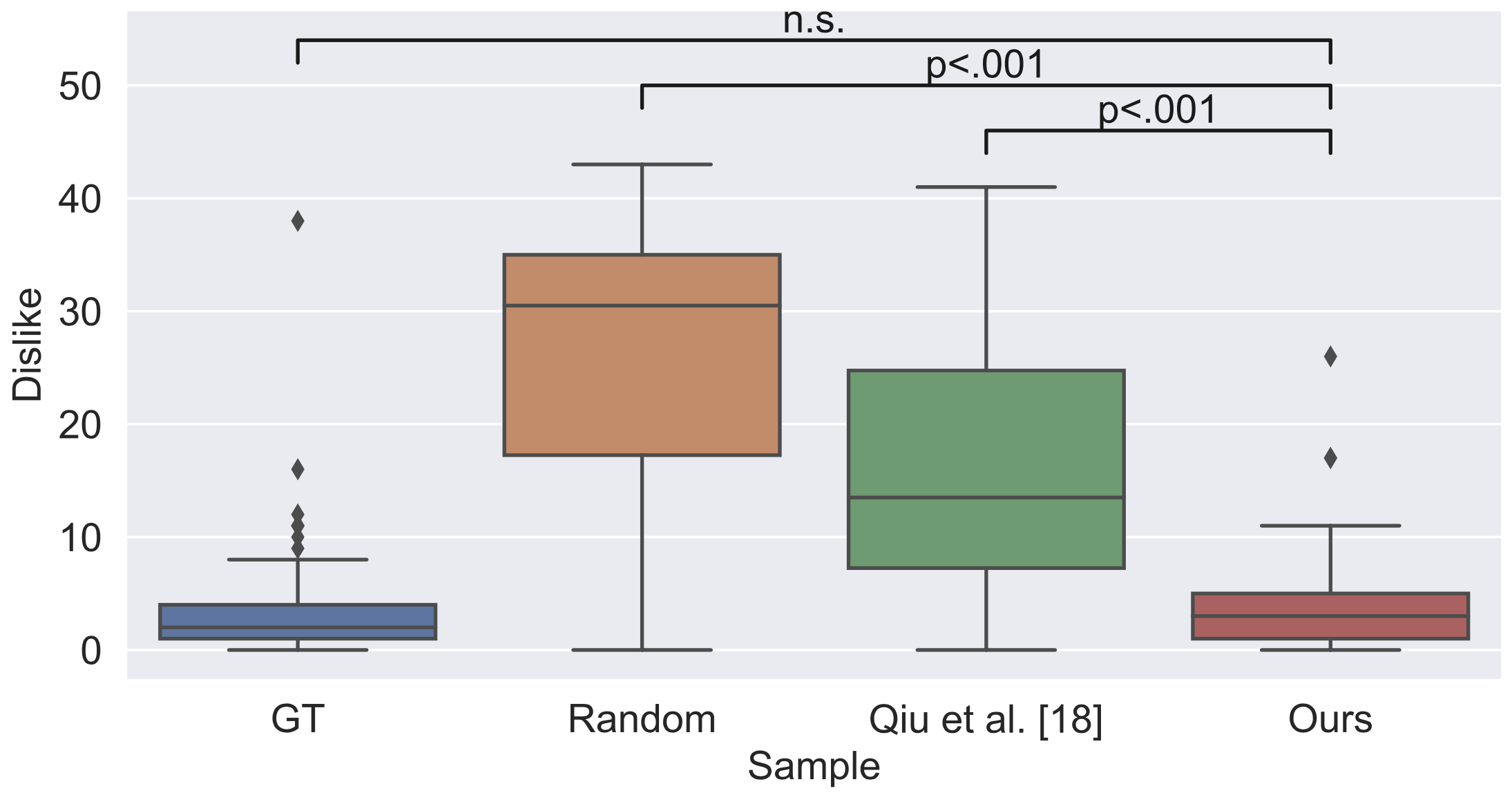}
    \caption{Evaluation results of bad design from designers.}
    \label{fig:eval_bad}
\end{figure}

\begin{figure}[t]
    \centering
    \includegraphics[width=1.0\linewidth]{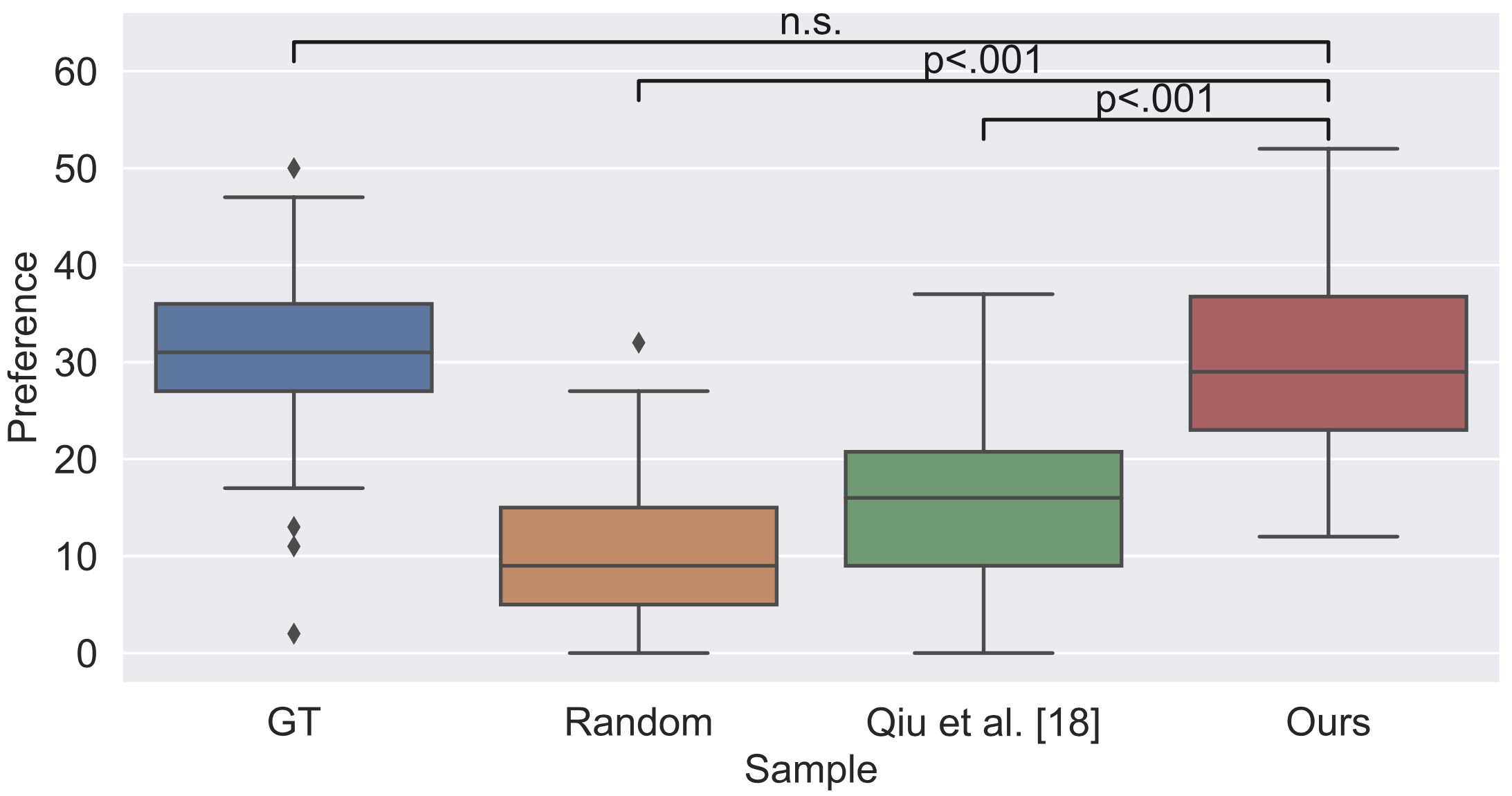}
    \caption{Evaluation results of good design from non-designers.}
    \label{fig:eval_good_non}
\end{figure}

\begin{figure}[t]
    \centering
    \includegraphics[width=1.0\linewidth]{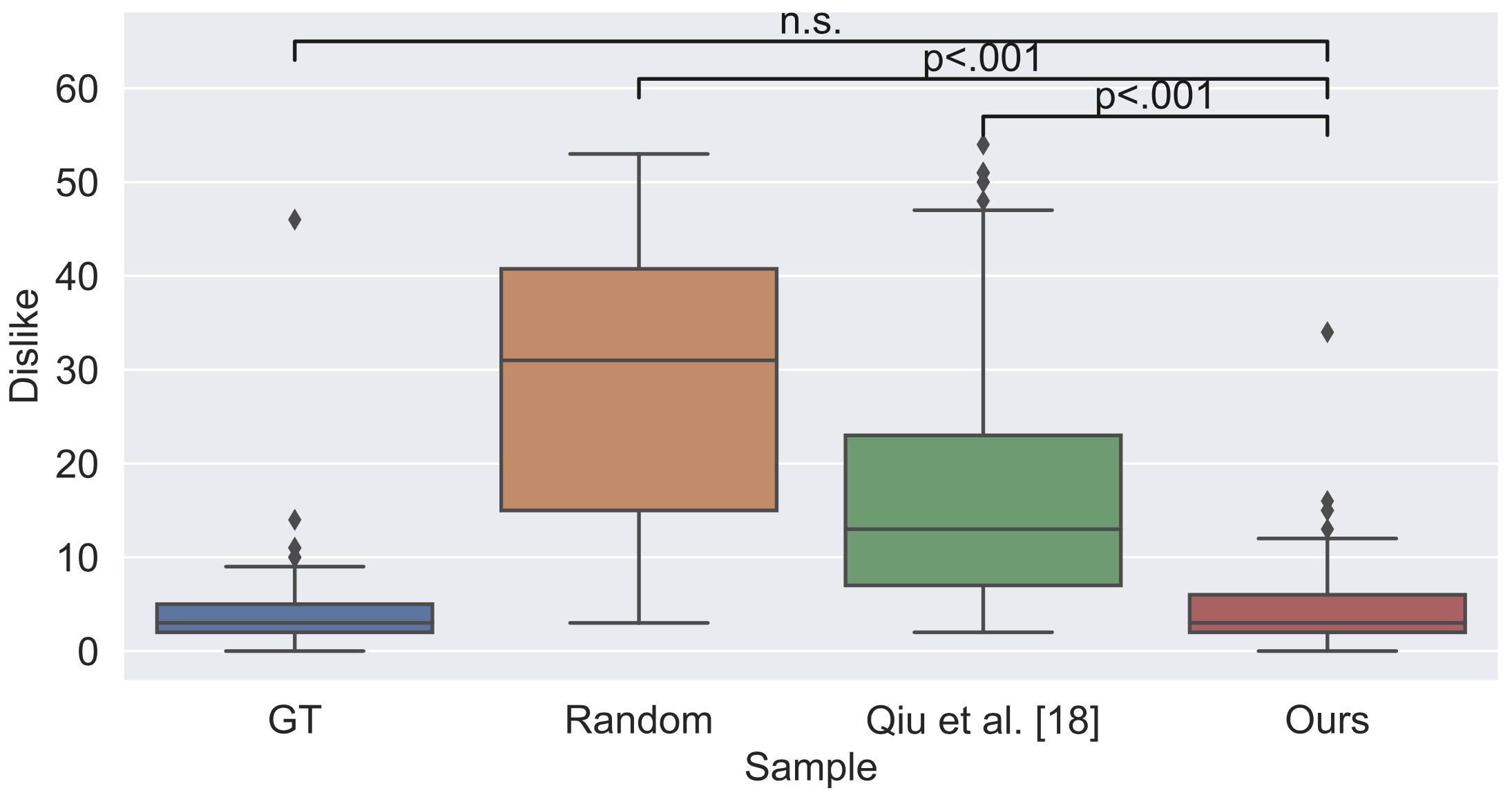}
    \caption{Evaluation results of bad design from non-designers.}
    \label{fig:eval_bad_non}
\end{figure}

We conduct a user study to investigate human perception on the outputs of our model. We select a specific number of colors and recolor the original design using recommended colors. We employ the palette-based photo recoloring method by Chang \textit{et al.}~\cite{chang2015palette} to obtain the recolored results. These recommendation results include one-color, two-color, and three-color recoloring cases as shown in Figure~\ref{fig:recolor_res}. We randomly selected 90 templates from the Crello-v2 test dataset, with 30 allocated for each recoloring case. In each case, colors are randomly selected from multiple palettes, associated with image, graphic, or text elements. We pick out the original design (GT), the recommended results from our proposed method, the related method by Qiu \textit{et al.}~\cite{qiu2023color}, and random color selection. These four design images are arranged together in random order in an evaluation question. The participants are asked to select any good designs and bad designs from four designs. We recruited 106 participants, consisting of 63 non-designers and 43 graphic designers.

The evaluation results for good and bad design selections by designers are presented in Figure~\ref{fig:eval_good} and Figure~\ref{fig:eval_bad}, while the results by non-designers are shown in Figure~\ref{fig:eval_good_non} and Figure~\ref{fig:eval_bad_non}. The results reveal that, for both designers and non-designers, the outcomes produced by our proposed method exhibit comparable performance to GT, and they demonstrate significantly higher preference and lower dislike compared to the related work and random selection (p < 0.001). Moreover, we conducted a detailed analysis of individual cases for one-color, two-color, or three-color recommendations. In this in-depth analysis, we observed a gradual enhancement in the preference value difference between our method and the related method, as the number of masked colors grew. We can speculate that the textual information can hold a more significant influence on color recommendation when less color-related information is presented.

Furthermore, we received positive feedback from four designers concerning the importance of selecting colors that align with textual impressions. They believed in the potential benefits of automated color recommendation for design purposes. However, they also pointed out that English text might lead to comprehension challenges for Japanese designers. This highlights the need for language localization and it should be considered in future research.

\subsection{Ablation studies and analysis}
\textbf{Impact of text-aware model and color order conditions.} Initially, we undertook ablation studies to systematically investigate the impact of the text-aware color model and color order conditions independently.
The ablation results are shown in Table~\ref{tab:model_order_res}. Our results indicate that the proposed text-aware color model outperforms the related color model without textual information for both area-based and lightness-based ordered color inputs. Moreover, the models trained with lightness-based ordered color inputs demonstrate superior performance compared to those trained with area-based ordered color inputs, as they encompass a more comprehensive color context.

\begin{table}[t] 
    \caption{Ablation study for text-aware color model and color order.} 
    \label{tab:model_order_res} 
    \setlength\tabcolsep{0pt} 
    \resizebox{\linewidth}{!}
    {
        \begin{tabular}{@{\extracolsep{5pt}} llcccccc}
        \hline
        \toprule
        Text & Color & \multicolumn{3}{c}{Accuracy@1$\uparrow$} & \multicolumn{3}{c}{Distribution@1$\uparrow$}\\
        \cmidrule{3-5} \cmidrule{6-8}
        Input & Order & {1 color} & {2 colors} & {3 colors} & {1 color} & {2 colors} & {3 colors}\\ 
        \midrule
        w/o & area & 36.72\% & 16.04\% & 6.45\% & 4.93 & 4.31 & 2.94 \\ 
        w/ & area & 39.06\% & 17.01\% & 7.52\% & 5.25 & 5.06 & 3.77 \\ 
        w/o & lightness & 44.53\% & 22.31\% & 12.69\% & 5.74 & 5.66 & 5.15 \\
        w/ & lightness & \textbf{47.13\%} & \textbf{26.22\%} & \textbf{15.67\%} & \textbf{6.04} & \textbf{6.64} & \textbf{6.30} \\
        \bottomrule
        \end{tabular}
    }
\end{table}

\textbf{Contributions of two cross-attention layers.} An experiment was carried out to examine the contributions of cross-attention (CA) and multi-cross-attention (MCA) layers within the multimodal masked color model. In this particular experiment, lightness-based ordered color inputs were utilized. The comparison results are shown in Table~\ref{tab:model_res}. The multimodal masked color model incorporating both cross-attention and multi-cross-attention layers yields marginally better performance.

\begin{table}[t] 
    \caption{Comparison results of two cross-attention layers.} 
    \label{tab:model_res} 
    \setlength\tabcolsep{0pt} 
    \resizebox{\linewidth}{!}
    {
        \begin{tabular}{@{\extracolsep{5pt}} llcccccc}
        \hline
        \toprule
        \multicolumn{2}{c}{Layers} & \multicolumn{3}{c}{Accuracy@1$\uparrow$} & \multicolumn{3}{c}{Distribution@1$\uparrow$}\\
        \cmidrule{1-2} \cmidrule{3-5} \cmidrule{6-8}
        CA & MCA & {1 color} & {2 colors} & {3 colors} & {1 color} & {2 colors} & {3 colors}\\ 
        \midrule
        w/ & w/o & 46.53\% & 25.27\% & 15.28\% & 5.91 & 6.44 & 5.95 \\
        w/o & w/ & 46.66\% & 25.64\% & 15.18\% & 6.03 & 6.49 & 6.06 \\
        w/ & w/ & \textbf{47.13\%} & \textbf{26.22\%} & \textbf{15.67\%} & \textbf{6.04} & \textbf{6.64} & \textbf{6.30} \\
        \bottomrule
        \end{tabular}
    }
\end{table}

\textbf{Comparison of CLIP and BERT models.} There are many popular LLMs available today. We compare the performance of the CLIP model with a text-only LLM, BERT~\cite{devlin2018bert} which has exhibited high performance on many NLP tasks. To compare the effectiveness of CLIP and BERT models for a color recommendation task, we obtained the text embeddings from these two models, and trained color models with our proposed method separately. The results in Table~\ref{tab:pretrained_model_res} indicate that the text embeddings from the CLIP model have greater performance than the text embeddings from the BERT model. In other words, multimodal pre-training models containing textural and visual information are more effective on multimodal color recommendation tasks than text-only pre-training models.

\begin{table}[t] 
    \caption{Comparison results of CLIP and BERT models.} 
    \label{tab:pretrained_model_res} 
    \setlength\tabcolsep{0pt} 
    \resizebox{\linewidth}{!}
    {
        \begin{tabular}{@{\extracolsep{8pt}} lcccccc}
        \hline
        \toprule
        & \multicolumn{3}{c}{Accuracy@1$\uparrow$} & \multicolumn{3}{c}{Distribution@1$\uparrow$}\\
        \cmidrule{2-4} \cmidrule{5-7}
        Pre-trained model & {1 color} & {2 colors} & {3 colors} & {1 color} & {2 colors} & {3 colors}\\ 
        \midrule
        BERT & 45.02\% & 22.55\% & 12.54\% & 5.59 & 6.09 & 5.30 \\ 
        CLIP & \textbf{47.13\%} & \textbf{26.22\%} & \textbf{15.67\%} & \textbf{6.04} & \textbf{6.64} & \textbf{6.30} \\
        \bottomrule
        \end{tabular}
    }
\end{table}

\begin{figure}[t]
    \centering
    \includegraphics[width=1.0\linewidth]{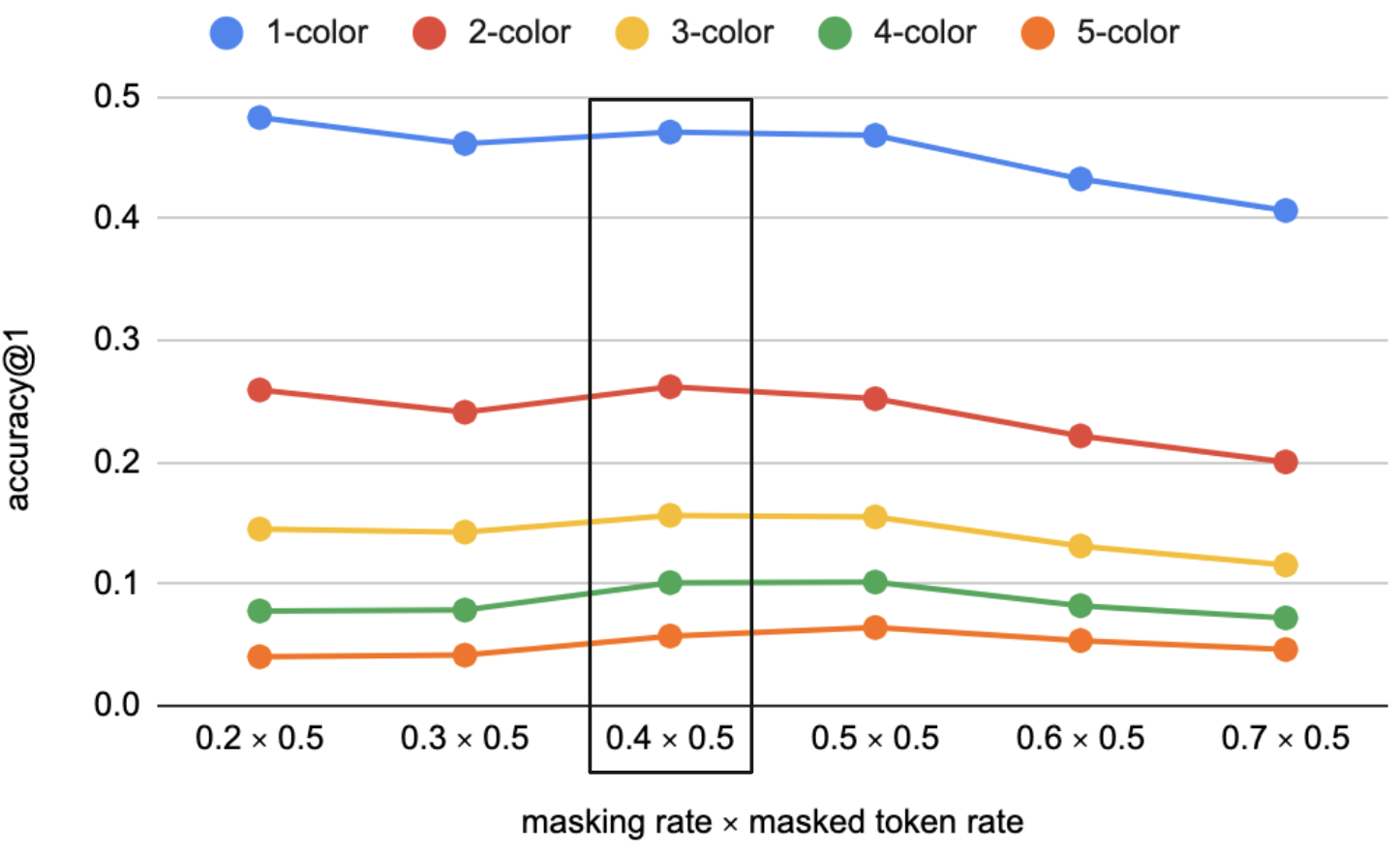}
    \caption{Accuracy@1 results of different masking rates.}
    \label{fig:mask_rate}
\end{figure}

\textbf{Optimal masking rate.} MLM conventionally masks {15\%} of tokens due to the belief that more masking would leave insufficient context to learn good representations \cite{devlin2018bert}. However, the optimal masking rate is not universally {15\%}, but should depend on kinds of factors \cite{wettig2022should}. We explore the optimal making rate and masked token rate for different data inputs. The prediction accuracy results for different masking rates in our proposed model are shown in Figure~\ref{fig:mask_rate}. Though a lower masking rate can yield good representations for the one-color recommendation, a {40\%} masking rate appears to be relatively optimal for the cases of more than one color prediction.

\section{Experiments of palette generation}
In this section, we present the application for full palette generation based on our proposed method. Full palette generation is a special case of text-aware color recommendation, referring to generating a complete color palette that corresponds to a given text. 
We compared our proposed model with Text-to-Palette Generation Networks (TPN) in the most relevant work~\cite{bahng2018coloring}. During our model training process, a masking rate of {80\%} and a masked token rate of {50\%} were chosen. On the other hand, TPN utilizing conditional GAN was trained for 500 epochs, following the implementation details provided in the related work.
For evaluation, we adopt the color diversity evaluation in the related work~\cite{bahng2018coloring} that calculates the average pairwise distance between the five colors within a palette. However, as color diversity alone cannot provide rigorous and comprehensive proof of a palette's superiority, we also measure the similarity between the generated palettes and the ground truth. We use the state-of-the-art palette similarity measurement, Dynamic Closest Color Warping (DCCW) method~\cite{kim2021dynamic}. DCCW calculates the minimum distance between colors in different palettes.

\subsection{Dataset}
We used the manually curated Palette-And-Text (PAT) dataset~\cite{bahng2018coloring} for the experiment. The PAT dataset contains the five-color palette and text pairs. The text description has strong associations with colors, that some words are direct color words (e.g., pink, grey) while others evoke a particular set of colors (e.g., summer, autumn). This dataset allows us to train our model for predicting a whole palette only with textural inputs. PAT dataset contains 10,183 text and five-color palette pairs. We randomly divide it into 8,147 / 1,018 / 1,018 data for training, validation, and testing.

\subsection{Performance comparison}
\begin{figure}[t]
    \centering
    \includegraphics[width=1.0\linewidth]{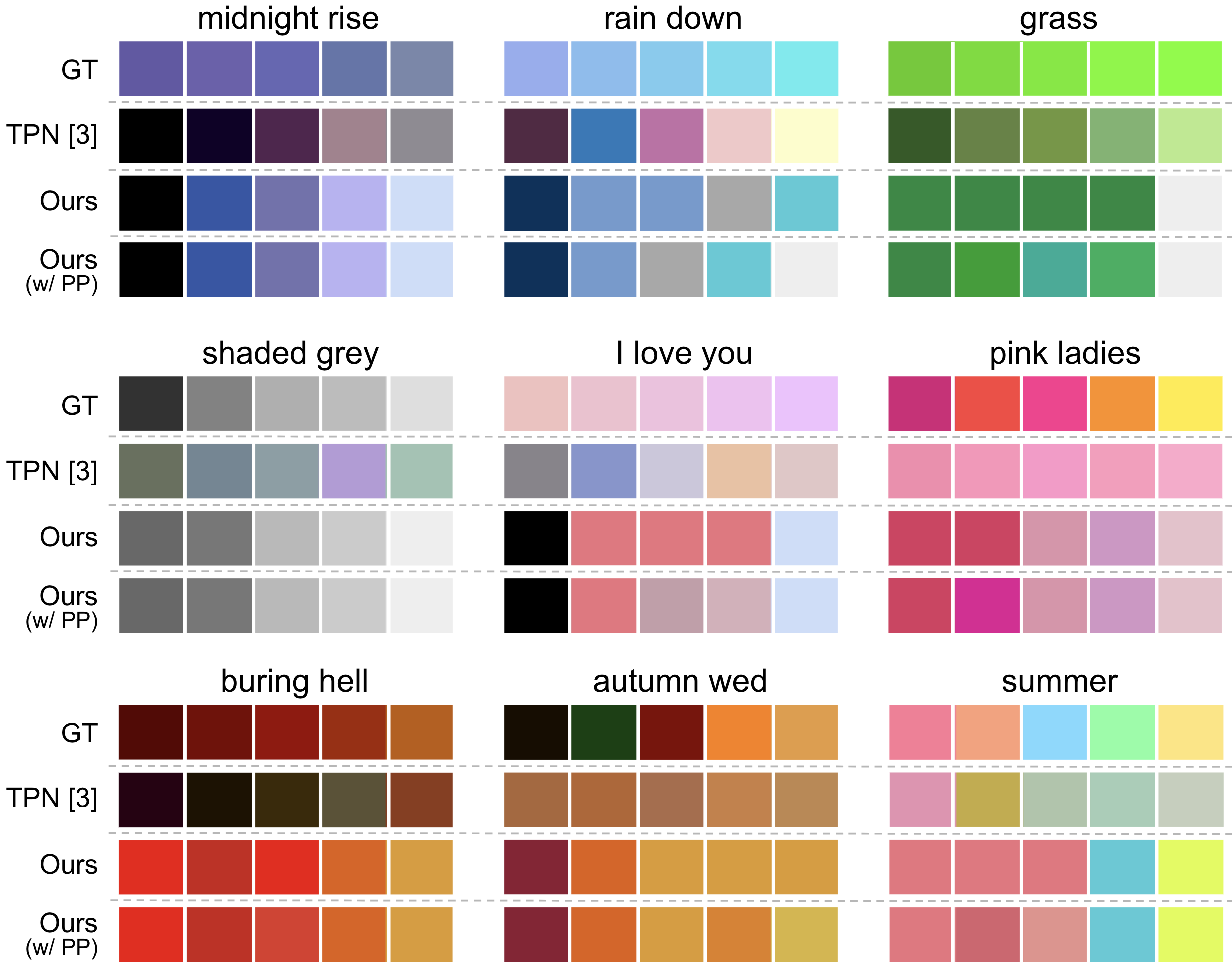}
    \caption{Qualitative analysis on textual context. We compare the generated palette results of our proposed method and the related work TPN~\cite{bahng2018coloring} with the ground truth.}
    \label{fig:generated_res}
\end{figure}

The comparison results of color diversity and palette similarity to GT in Table~\ref{tab:div_sim_res} indicated that our generated palettes have higher diversity and closer similarity to GT.
Moreover, we output a series of palettes based on various textual contexts. A comparison between generated palettes of our proposed method and the related work TPN with GT can be observed in Figure~\ref{fig:generated_res}. It is noted that higher diversity may not necessarily be a critical factor, as incorporating colors in the palette that are irrelevant to the text would increase diversity. On the other hand, palette similarity to GT holds greater importance, as it indicates whether the recommended results contain key colors that accurately convey the intended semantics.

In some cases, the same color may be recommended for different masked positions with the highest probability. For example, the first four colors in our results for `grass' in Figure~\ref{fig:generated_res} are the same. To address this issue, we introduce a post-processing (PP) step to eliminate duplicated colors in a palette. We choose a new color with the next highest probability if a recommended color is the same as a previous color in the sequence, continuing until a non-duplicated color is selected. As demonstrated in Table~\ref{tab:div_sim_res}, the PP procedure increased color diversity and improved the palette similarity to GT.

\begin{table}[t] 
    \caption{Quantitative analysis results of color diversity and palette similarity to GT.} 
    \label{tab:div_sim_res} 
    \setlength\tabcolsep{0pt} 
    \resizebox{\linewidth}{!}
    {
        \begin{tabular}{@{\extracolsep{8pt}} llcccc}
        \hline
        \toprule
        & & \multicolumn{2}{c}{Diversity@1$\uparrow$} & \multicolumn{2}{c}{Similarity to GT$\downarrow$}\\
        \cmidrule{3-4} \cmidrule{5-6}
        Generated palettes & PP & {Mean} & {Std} & {Mean} & {Std} \\ 
        \midrule
        TPN~\cite{bahng2018coloring} & w/o & 22.21 & 10.78 & 29.26 & 13.35 \\ 
        Ours & w/o & \textbf{29.92} & 10.27 & \textbf{28.14} & 12.91 \\
        Ours & w & \textbf{33.33} & 9.38 & \textbf{27.06} & 12.38 \\
        GT & - & 26.17 & 13.84 & - & - \\
        \bottomrule
        \end{tabular}
    }
\end{table}

\section{Conclusions}
In this paper, we presented a multimodal masked color model for vector graphic documents, with reordering the color input based on lightness in CIELAB color space and CLIP-based text representation. Through quantitative and qualitative evaluations, our method is improved to have greater performance on the accuracy, color distribution, and user experience compared to prior methods in recommending colors based on the given colors and textual contexts. Moreover, our proposal is applicable for full palette generation and surpasses related work on color diversity and palette similarity to the ground truth.

In this research, we employed accuracy and palette similarity to the ground truth as the primary metrics for evaluating the effectiveness of our recommended colors, aiming to achieve results that are closely aligned with the actual designs. However, the current evaluation measures do not account for other aspects, such as aesthetics and textual relevance. In future work, we plan to explore more comprehensive evaluation metrics for color recommendation.


\bibliographystyle{ACM-Reference-Format}
\bibliography{base}


\begin{thebibliography}{26}


\ifx \showCODEN    \undefined \def \showCODEN     #1{\unskip}     \fi
\ifx \showDOI      \undefined \def \showDOI       #1{#1}\fi
\ifx \showISBNx    \undefined \def \showISBNx     #1{\unskip}     \fi
\ifx \showISBNxiii \undefined \def \showISBNxiii  #1{\unskip}     \fi
\ifx \showISSN     \undefined \def \showISSN      #1{\unskip}     \fi
\ifx \showLCCN     \undefined \def \showLCCN      #1{\unskip}     \fi
\ifx \shownote     \undefined \def \shownote      #1{#1}          \fi
\ifx \showarticletitle \undefined \def \showarticletitle #1{#1}   \fi
\ifx \showURL      \undefined \def \showURL       {\relax}        \fi
\providecommand\bibfield[2]{#2}
\providecommand\bibinfo[2]{#2}
\providecommand\natexlab[1]{#1}
\providecommand\showeprint[2][]{arXiv:#2}

\bibitem[Ado(4 30)]%
        {AdobeColor}
 \bibinfo{year}{Accessed 2023-4-30}\natexlab{}.
\newblock \bibinfo{title}{Adobe Color}.
\newblock \bibinfo{howpublished}{\url{https://color.adobe.com/}}.
\newblock


\bibitem[Col(4 30)]%
        {ColourLovers}
 \bibinfo{year}{Accessed 2023-4-30}\natexlab{}.
\newblock \bibinfo{title}{COLOURLovers}.
\newblock \bibinfo{howpublished}{\url{https://www.colourlovers.com/}}.
\newblock


\bibitem[Bahng et~al\mbox{.}(2018)]%
        {bahng2018coloring}
\bibfield{author}{\bibinfo{person}{Hyojin Bahng}, \bibinfo{person}{Seungjoo
  Yoo}, \bibinfo{person}{Wonwoong Cho}, \bibinfo{person}{David~Keetae Park},
  \bibinfo{person}{Ziming Wu}, \bibinfo{person}{Xiaojuan Ma}, {and}
  \bibinfo{person}{Jaegul Choo}.} \bibinfo{year}{2018}\natexlab{}.
\newblock \showarticletitle{Coloring with words: Guiding image colorization
  through text-based palette generation}. In
  \bibinfo{booktitle}{\emph{Proceedings of the european conference on computer
  vision (eccv)}}. \bibinfo{pages}{431--447}.
\newblock


\bibitem[Chang et~al\mbox{.}(2015)]%
        {chang2015palette}
\bibfield{author}{\bibinfo{person}{Huiwen Chang}, \bibinfo{person}{Ohad Fried},
  \bibinfo{person}{Yiming Liu}, \bibinfo{person}{Stephen DiVerdi}, {and}
  \bibinfo{person}{Adam Finkelstein}.} \bibinfo{year}{2015}\natexlab{}.
\newblock \showarticletitle{Palette-based photo recoloring.}
\newblock \bibinfo{journal}{\emph{ACM Trans. Graph.}} \bibinfo{volume}{34},
  \bibinfo{number}{4} (\bibinfo{year}{2015}), \bibinfo{pages}{139--1}.
\newblock


\bibitem[Chijiiwa(1987)]%
        {chijiiwa1987color}
\bibfield{author}{\bibinfo{person}{Hideaki Chijiiwa}.}
  \bibinfo{year}{1987}\natexlab{}.
\newblock \bibinfo{booktitle}{\emph{Color harmony: a guide to creative color
  combinations}}. Vol.~\bibinfo{volume}{1}.
\newblock \bibinfo{publisher}{Rockport Pub}.
\newblock


\bibitem[Cho et~al\mbox{.}(2017)]%
        {cho2017palettenet}
\bibfield{author}{\bibinfo{person}{Junho Cho}, \bibinfo{person}{Sangdoo Yun},
  \bibinfo{person}{Kyoung Mu~Lee}, {and} \bibinfo{person}{Jin Young~Choi}.}
  \bibinfo{year}{2017}\natexlab{}.
\newblock \showarticletitle{Palettenet: Image recolorization with given color
  palette}. In \bibinfo{booktitle}{\emph{Proceedings of the ieee conference on
  computer vision and pattern recognition workshops}}. \bibinfo{pages}{62--70}.
\newblock


\bibitem[Devlin et~al\mbox{.}(2018)]%
        {devlin2018bert}
\bibfield{author}{\bibinfo{person}{Jacob Devlin}, \bibinfo{person}{Ming-Wei
  Chang}, \bibinfo{person}{Kenton Lee}, {and} \bibinfo{person}{Kristina
  Toutanova}.} \bibinfo{year}{2018}\natexlab{}.
\newblock \showarticletitle{Bert: Pre-training of deep bidirectional
  transformers for language understanding}.
\newblock \bibinfo{journal}{\emph{arXiv preprint arXiv:1810.04805}}
  (\bibinfo{year}{2018}).
\newblock


\bibitem[Jahanian et~al\mbox{.}(2013)]%
        {jahanian2013recommendation}
\bibfield{author}{\bibinfo{person}{Ali Jahanian}, \bibinfo{person}{Jerry Liu},
  \bibinfo{person}{Qian Lin}, \bibinfo{person}{Daniel Tretter},
  \bibinfo{person}{Eamonn O'Brien-Strain}, \bibinfo{person}{Seungyon~Claire
  Lee}, \bibinfo{person}{Nic Lyons}, {and} \bibinfo{person}{Jan Allebach}.}
  \bibinfo{year}{2013}\natexlab{}.
\newblock \showarticletitle{Recommendation system for automatic design of
  magazine covers}. In \bibinfo{booktitle}{\emph{Proceedings of the 2013
  international conference on Intelligent user interfaces}}.
  \bibinfo{pages}{95--106}.
\newblock


\bibitem[Kawakami et~al\mbox{.}(2016)]%
        {kawakami2016character}
\bibfield{author}{\bibinfo{person}{Kazuya Kawakami}, \bibinfo{person}{Chris
  Dyer}, \bibinfo{person}{Bryan~R Routledge}, {and} \bibinfo{person}{Noah~A
  Smith}.} \bibinfo{year}{2016}\natexlab{}.
\newblock \showarticletitle{Character Sequence Models for Colorful Words}. In
  \bibinfo{booktitle}{\emph{Proceedings of the 2016 Conference on Empirical
  Methods in Natural Language Processing}}. \bibinfo{pages}{1949--1954}.
\newblock


\bibitem[Kikuchi et~al\mbox{.}(2023)]%
        {kikuchi2023generative}
\bibfield{author}{\bibinfo{person}{Kotaro Kikuchi}, \bibinfo{person}{Naoto
  Inoue}, \bibinfo{person}{Mayu Otani}, \bibinfo{person}{Edgar Simo-Serra},
  {and} \bibinfo{person}{Kota Yamaguchi}.} \bibinfo{year}{2023}\natexlab{}.
\newblock \showarticletitle{Generative Colorization of Structured Mobile Web
  Pages}. In \bibinfo{booktitle}{\emph{Proceedings of the IEEE/CVF Winter
  Conference on Applications of Computer Vision}}. \bibinfo{pages}{3650--3659}.
\newblock


\bibitem[Kim et~al\mbox{.}(2022)]%
        {kim2022colorbo}
\bibfield{author}{\bibinfo{person}{Eunseo Kim}, \bibinfo{person}{Jeongmin
  Hong}, \bibinfo{person}{Hyuna Lee}, {and} \bibinfo{person}{Minsam Ko}.}
  \bibinfo{year}{2022}\natexlab{}.
\newblock \showarticletitle{Colorbo: Envisioned Mandala Coloringthrough
  Human-AI Collaboration}. In \bibinfo{booktitle}{\emph{27th International
  Conference on Intelligent User Interfaces}}. \bibinfo{pages}{15--26}.
\newblock


\bibitem[Kim and Choi(2021)]%
        {kim2021dynamic}
\bibfield{author}{\bibinfo{person}{Suzi Kim} {and} \bibinfo{person}{Sunghee
  Choi}.} \bibinfo{year}{2021}\natexlab{}.
\newblock \showarticletitle{Dynamic closest color warping to sort and compare
  palettes}.
\newblock \bibinfo{journal}{\emph{ACM Transactions on Graphics (TOG)}}
  \bibinfo{volume}{40}, \bibinfo{number}{4} (\bibinfo{year}{2021}),
  \bibinfo{pages}{1--15}.
\newblock


\bibitem[Kita and Miyata(2016)]%
        {kita2016aesthetic}
\bibfield{author}{\bibinfo{person}{Naoki Kita} {and} \bibinfo{person}{Kazunori
  Miyata}.} \bibinfo{year}{2016}\natexlab{}.
\newblock \showarticletitle{Aesthetic rating and color suggestion for color
  palettes}. In \bibinfo{booktitle}{\emph{Computer Graphics Forum}},
  Vol.~\bibinfo{volume}{35}. Wiley Online Library, \bibinfo{pages}{127--136}.
\newblock


\bibitem[Maheshwari et~al\mbox{.}(2021)]%
        {maheshwari2021generating}
\bibfield{author}{\bibinfo{person}{Paridhi Maheshwari}, \bibinfo{person}{Nihal
  Jain}, \bibinfo{person}{Praneetha Vaddamanu}, \bibinfo{person}{Dhananjay
  Raut}, \bibinfo{person}{Shraiysh Vaishay}, {and} \bibinfo{person}{Vishwa
  Vinay}.} \bibinfo{year}{2021}\natexlab{}.
\newblock \showarticletitle{Generating Compositional Color Representations from
  Text}. In \bibinfo{booktitle}{\emph{Proceedings of the 30th ACM International
  Conference on Information \& Knowledge Management}}.
  \bibinfo{pages}{1222--1231}.
\newblock


\bibitem[Monroe et~al\mbox{.}(2017)]%
        {monroe2017colors}
\bibfield{author}{\bibinfo{person}{Will Monroe}, \bibinfo{person}{Robert~XD
  Hawkins}, \bibinfo{person}{Noah~D Goodman}, {and}
  \bibinfo{person}{Christopher Potts}.} \bibinfo{year}{2017}\natexlab{}.
\newblock \showarticletitle{Colors in context: A pragmatic neural model for
  grounded language understanding}.
\newblock \bibinfo{journal}{\emph{Transactions of the Association for
  Computational Linguistics}}  \bibinfo{volume}{5} (\bibinfo{year}{2017}),
  \bibinfo{pages}{325--338}.
\newblock


\bibitem[O'Donovan et~al\mbox{.}(2011)]%
        {o2011color}
\bibfield{author}{\bibinfo{person}{Peter O'Donovan}, \bibinfo{person}{Aseem
  Agarwala}, {and} \bibinfo{person}{Aaron Hertzmann}.}
  \bibinfo{year}{2011}\natexlab{}.
\newblock \showarticletitle{Color compatibility from large datasets}.
\newblock In \bibinfo{booktitle}{\emph{ACM SIGGRAPH 2011 papers}}.
  \bibinfo{pages}{1--12}.
\newblock


\bibitem[Qiu et~al\mbox{.}(2022)]%
        {qiu2022intelligent}
\bibfield{author}{\bibinfo{person}{Qianru Qiu}, \bibinfo{person}{Mayu Otani},
  {and} \bibinfo{person}{Yuki Iwazaki}.} \bibinfo{year}{2022}\natexlab{}.
\newblock \showarticletitle{An Intelligent Color Recommendation Tool for
  Landing Page Design}. In \bibinfo{booktitle}{\emph{27th International
  Conference on Intelligent User Interfaces}}. \bibinfo{pages}{26--29}.
\newblock


\bibitem[Qiu et~al\mbox{.}(2023)]%
        {qiu2023color}
\bibfield{author}{\bibinfo{person}{Qianru Qiu}, \bibinfo{person}{Xueting Wang},
  \bibinfo{person}{Mayu Otani}, {and} \bibinfo{person}{Yuki Iwazaki}.}
  \bibinfo{year}{2023}\natexlab{}.
\newblock \showarticletitle{Color Recommendation for Vector Graphic Documents
  based on Multi-Palette Representation}. In
  \bibinfo{booktitle}{\emph{Proceedings of the IEEE/CVF Winter Conference on
  Applications of Computer Vision}}. \bibinfo{pages}{3621--3629}.
\newblock


\bibitem[Radford et~al\mbox{.}(2021)]%
        {radford2021learning}
\bibfield{author}{\bibinfo{person}{Alec Radford}, \bibinfo{person}{Jong~Wook
  Kim}, \bibinfo{person}{Chris Hallacy}, \bibinfo{person}{Aditya Ramesh},
  \bibinfo{person}{Gabriel Goh}, \bibinfo{person}{Sandhini Agarwal},
  \bibinfo{person}{Girish Sastry}, \bibinfo{person}{Amanda Askell},
  \bibinfo{person}{Pamela Mishkin}, \bibinfo{person}{Jack Clark},
  {et~al\mbox{.}}} \bibinfo{year}{2021}\natexlab{}.
\newblock \showarticletitle{Learning transferable visual models from natural
  language supervision}. In \bibinfo{booktitle}{\emph{International conference
  on machine learning}}. PMLR, \bibinfo{pages}{8748--8763}.
\newblock


\bibitem[Vaswani et~al\mbox{.}(2017)]%
        {vaswani2017attention}
\bibfield{author}{\bibinfo{person}{Ashish Vaswani}, \bibinfo{person}{Noam
  Shazeer}, \bibinfo{person}{Niki Parmar}, \bibinfo{person}{Jakob Uszkoreit},
  \bibinfo{person}{Llion Jones}, \bibinfo{person}{Aidan~N Gomez},
  \bibinfo{person}{{\L}ukasz Kaiser}, {and} \bibinfo{person}{Illia
  Polosukhin}.} \bibinfo{year}{2017}\natexlab{}.
\newblock \showarticletitle{Attention is all you need}.
\newblock \bibinfo{journal}{\emph{Advances in neural information processing
  systems}}  \bibinfo{volume}{30} (\bibinfo{year}{2017}).
\newblock


\bibitem[Wettig et~al\mbox{.}(2022)]%
        {wettig2022should}
\bibfield{author}{\bibinfo{person}{Alexander Wettig}, \bibinfo{person}{Tianyu
  Gao}, \bibinfo{person}{Zexuan Zhong}, {and} \bibinfo{person}{Danqi Chen}.}
  \bibinfo{year}{2022}\natexlab{}.
\newblock \showarticletitle{Should you mask 15\% in masked language modeling?}
\newblock \bibinfo{journal}{\emph{arXiv preprint arXiv:2202.08005}}
  (\bibinfo{year}{2022}).
\newblock


\bibitem[Yamaguchi(2021)]%
        {yamaguchi2021canvasvae}
\bibfield{author}{\bibinfo{person}{Kota Yamaguchi}.}
  \bibinfo{year}{2021}\natexlab{}.
\newblock \showarticletitle{Canvasvae: learning to generate vector graphic
  documents}. In \bibinfo{booktitle}{\emph{Proceedings of the IEEE/CVF
  International Conference on Computer Vision}}. \bibinfo{pages}{5481--5489}.
\newblock


\bibitem[Yang et~al\mbox{.}(2016)]%
        {yang2016automatic}
\bibfield{author}{\bibinfo{person}{Xuyong Yang}, \bibinfo{person}{Tao Mei},
  \bibinfo{person}{Ying-Qing Xu}, \bibinfo{person}{Yong Rui}, {and}
  \bibinfo{person}{Shipeng Li}.} \bibinfo{year}{2016}\natexlab{}.
\newblock \showarticletitle{Automatic generation of visual-textual presentation
  layout}.
\newblock \bibinfo{journal}{\emph{ACM Transactions on Multimedia Computing,
  Communications, and Applications (TOMM)}} \bibinfo{volume}{12},
  \bibinfo{number}{2} (\bibinfo{year}{2016}), \bibinfo{pages}{1--22}.
\newblock


\bibitem[Yuan et~al\mbox{.}(2021)]%
        {yuan2021infocolorizer}
\bibfield{author}{\bibinfo{person}{Lin-Ping Yuan}, \bibinfo{person}{Ziqi Zhou},
  \bibinfo{person}{Jian Zhao}, \bibinfo{person}{Yiqiu Guo},
  \bibinfo{person}{Fan Du}, {and} \bibinfo{person}{Huamin Qu}.}
  \bibinfo{year}{2021}\natexlab{}.
\newblock \showarticletitle{Infocolorizer: Interactive recommendation of color
  palettes for infographics}.
\newblock \bibinfo{journal}{\emph{IEEE Transactions on Visualization and
  Computer Graphics}} \bibinfo{volume}{28}, \bibinfo{number}{12}
  (\bibinfo{year}{2021}), \bibinfo{pages}{4252--4266}.
\newblock


\bibitem[Zaccarin and Liu(1993)]%
        {zaccarin1993novel}
\bibfield{author}{\bibinfo{person}{Andr{\'e} Zaccarin} {and}
  \bibinfo{person}{Bede Liu}.} \bibinfo{year}{1993}\natexlab{}.
\newblock \showarticletitle{A novel approach for coding color quantized
  images}.
\newblock \bibinfo{journal}{\emph{IEEE Transactions on Image Processing}}
  \bibinfo{volume}{2}, \bibinfo{number}{4} (\bibinfo{year}{1993}),
  \bibinfo{pages}{442--453}.
\newblock


\bibitem[Zhang et~al\mbox{.}(2017)]%
        {zhang2017palette}
\bibfield{author}{\bibinfo{person}{Qing Zhang}, \bibinfo{person}{Chunxia Xiao},
  \bibinfo{person}{Hanqiu Sun}, {and} \bibinfo{person}{Feng Tang}.}
  \bibinfo{year}{2017}\natexlab{}.
\newblock \showarticletitle{Palette-based image recoloring using color
  decomposition optimization}.
\newblock \bibinfo{journal}{\emph{IEEE Transactions on Image Processing}}
  \bibinfo{volume}{26}, \bibinfo{number}{4} (\bibinfo{year}{2017}),
  \bibinfo{pages}{1952--1964}.
\newblock


\end{thebibliography}

\end{document}